%% file: tmi.tex
\documentclass[journal,twoside,web]{ieeecolor}
\usepackage{tmi}
\usepackage{cite}
\usepackage{amsmath,amssymb,amsfonts}
\usepackage{algorithmic}
\usepackage{graphicx}
\usepackage{textcomp}

\usepackage{hyperref}
\usepackage{bm}
\hypersetup{hypertex=true,
colorlinks=true,
linkcolor=blue,
anchorcolor=blue,
citecolor=blue}
\usepackage{multicol}
\usepackage{multirow}
\usepackage{booktabs}
\usepackage[table]{xcolor}
\definecolor{deepgreen}{RGB}{0,180,0}
\definecolor{lightblue}{RGB}{236, 250, 254}
\usepackage{graphicx}
\usepackage{textcomp}
\usepackage{bbding}
\usepackage{svg}
\usepackage{pifont}
\usepackage{helvet}

\def\BibTeX{{\rm B\kern-.05em{\sc i\kern-.025em b}\kern-.08em
    T\kern-.1667em\lower.7ex\hbox{E}\kern-.125emX}}
\begin{document}

\title{VeCAS: Vessel-Focused Contrast-Free Angiogram Synthesis for Vascular Interventions}

\author{
De-Xing Huang,
Chen-Yu Wang, 
Hao Liang, 
Xiao-Hu Zhou,  
Mei-Jiang Gui, 
Tian-Yu Xiang, 
Qin-Yi Zhang, \\
Chen Wang, 
Xiao-Liang Xie, 
Shi-Qi Liu,  
Ming-Yuan Liu,  
Zhen-Chang Wang, and Zeng-Guang Hou
\thanks{This work was supported in part by the National Natural Science Foundation of China under Grant 62373351, Grant 62503474, Grant 82327801, Grant 62303463, Grant 52475038, in part by the Chinese Academy of Sciences Project for Research Instrument Development under Grant No. PTYQ2026YZ0072, Young Scientists in Basic Research under Grant No. YSBR-104, in part by the Beijing Natural Science Foundation under Grant F252068, Grant 4254107, in part by Beijing Nova Program under Grant 20250484813, in part by China Postdoctoral Science Foundation under Grant 2024M763535, in part by the Postdoctoral Fellowship Program of CPSF under Grant GZC20251170, and in part by Xiaomi Young Talents Program/Xiaomi Foundation.}
\thanks{$\dagger$: D.-X. Huang, C.-Y. Wang, and H. Liang contributed equally to this work (email: \href{mailto:huangdexing2022@ia.ac.cn}{\tt huangdexing2022@ia.ac.cn}, \href{mailto:chenyuwang.cs@gmail.com}{\tt chenyuwang.cs@gmail.com}, \href{mailto:lh250312@163.com}{\tt lh250312@163.com}).}
\thanks{$\ast$ Corresponding authors: Xiao-Hu Zhou, Ming-Yuan Liu, and Zeng-Guang Hou (email: {\tt \{\href{mailto:xiaohu.zhou@ia.ac.cn}{xiaohu.zhou}, \href{mailto:zengguang.hou@ia.ac.cn}{zengguang.hou}\}@ia.ac.cn}, {\tt \href{mailto:dr.mingyuanliu@pku.edu.cn}{dr.mingyuanliu@pku.edu.cn}}).}
\thanks{D.-X. Huang, C.-Y. Wang, X.-H. Zhou, M.-J. Gui, T.-Y. Xiang, Q.-Y. Zhang, C. Wang, X.-L. Xie, S.-Q. Liu, and Z.-G. Hou are with the State Key Laboratory of Multimodal Artificial Intelligence Systems, Institute of Automation, Chinese Academy of Sciences, Beijing 100190, China.}
\thanks{D.-X. Huang, X.-H. Zhou, T.-Y. Xiang, Q.-Y. Zhang, C. Wang, and Z.-G. Hou are also with the School of Artificial Intelligence, University of Chinese Academy of Sciences, Beijing 100049, China.}
\thanks{C.-Y. Wang is also with the School of Advanced Interdisciplinary Sciences, University of Chinese Academy of Sciences, Beijing 101408, China.}
\thanks{H. Liang, M.-Y. Liu, and Z.-C. Wang are with the Beijing Friendship Hospital, Capital Medical University, Beijing 100050, China.}
}

\maketitle

\begin{abstract}
X-ray angiography relies on iodinated contrast agents to visualize vascular structures during image-guided interventions. However, contrast administration carries risks of adverse events, motivating the development of contrast-free alternatives. Generating X-ray angiograms directly from non-contrast X-ray images offers a potential solution, but existing approaches remain limited by \textit{(i) insufficient control over vascular localization} and \textit{(ii) inefficient modeling of redundant background content}. To address these challenges, we propose \textbf{VeCAS}, a two-stage \underline{Ve}ssel-focused \underline{C}ontrast-free \underline{A}ngiogram \underline{S}ynthesis framework that separates vascular structure localization from angiographic appearance synthesis. In Stage I, a discriminative model localizes vascular structures in non-contrast X-ray images, while cross-modality latent distillation transfers vessel-sensitive knowledge from X-ray angiograms during training. In Stage II, a vessel-focused inpainting model synthesizes angiographic appearance within the localized vascular regions while preserving the non-vascular background. Experiments on an in-house lower-limb vascular intervention dataset show that VeCAS outperforms the comparison methods in terms of vascular structural fidelity and image quality. Visual Turing tests and physician assessments indicate the perceptual realism of the synthesized angiograms. In addition, robotic guidewire navigation experiments in vascular phantoms show that VeCAS guidance reduces the time to target by 41.4\% and the number of operation steps by 40.7\% compared with non-contrast guidance. Together, these results suggest the potential of VeCAS to serve as ``\textit{meta contrast agent}'' for vascular interventions.
\end{abstract}


\section{Introduction} \label{sec:introduction}
\input{Section1_Introduction}

\section{Related Work} \label{sec:related_work}
\input{Section2_Related_Work}

\section{Method} \label{sec:method}
\input{Section3_Method}

\section{Results} \label{sec:results}
\input{Section4_Results}


\section{Conclusion} \label{sec:conclusion}
\input{Section6_Conclusion}

\bibliographystyle{IEEEtran}
\bibliography{reference}

\end{document}

%% file: Section1_Introduction.tex
\IEEEPARstart{I}{mage}-guided vascular interventions have become an important therapeutic option for peripheral artery disease because of their minimally invasive nature, faster postoperative recovery, and lower overall risk than conventional surgical procedures~\cite{thukkani2015endovascular, hiramoto2018interventions}. During interventions, X-ray angiography provides real-time $2$D imaging, enabling physicians to observe the spatial relationship between vascular structures and interventional instruments, such as guidewires, catheters, and balloons~\cite{zhao2025large,zhao2026generative}. Since vessels are poorly visible in non-contrast X-ray images, iodinated contrast agents are routinely injected to enhance vascular visibility, as shown in Fig.~\ref{fig:motivation}(Top).

Despite its clinical importance, the use of iodinated contrast agents is associated with several potential risks, including life-threatening allergic reactions~\cite{clement2018immediate} and acute kidney injury (AKI)~\cite{fahling2017understanding}. Although these adverse events are multifactorial, the administered volume of iodinated contrast agents has been widely recognized as an important contributing factor~\cite{mehran2006contrast,tehrani2013contrast,giacoppo2015impact,bartholomew2004impact}. A recent cross-sectional study involving $1.3$ million patients undergoing vascular interventions reported that each additional $75$ mL of contrast agent was associated with a $42\%$ increase in the risk of AKI~\cite{amin2017association}. These findings have motivated increasing interest in ultra-low-contrast and non-contrast vascular interventions, which aim to reduce contrast-related risks while preserving effective image guidance~\cite{wang2025angio,huang2025cas}. However, without contrast enhancement, vascular structures are difficult to observe, making instrument navigation more challenging and potentially reducing procedural efficiency.

\input{Figure/Figure_Motivation}

To reduce the reliance on iodinated contrast agents, several strategies have been explored in clinical practice, including intravascular ultrasound- or physiology-guided interventions~\cite{ali2016imaging,shibata2022feasibility}, roadmap-based guidance~\cite{hennessey2024dynamic}, and ${\rm CO}_2$ angiography~\cite{lee2023carbon}. However, these approaches usually require intravascular imaging, alternative contrast agents, or previously acquired angiographic information. Direct angiographic visualization from non-contrast X-ray images remains underexplored. Recent advances in artificial intelligence-generated content (AIGC) have demonstrated the ability to generate realistic visual content~\cite{rombach2022high,brooks2024video}. This progress raises the question: \textit{Can AI serve as ``\underline{meta contrast agent}'' to synthesize X-ray angiograms that approximate the vascular visualization provided by actual contrast agent?} As illustrated in Fig.~\ref{fig:motivation}(Bottom), such a capability could provide a potential contrast-free strategy for vascular interventions.

Nevertheless, developing a reliable contrast-free angiogram synthesis method remains challenging. Existing approaches mainly suffer from two limitations: \textbf{\textit{(i)} Insufficient control over vascular localization.} Most existing methods formulate this task as image-to-image translation~\cite{isola2017image,zhu2017unpaired} and learn a global mapping from the source domain to the target domain~\cite{huang2025cas}. Consequently, these methods tend to emphasize the overall realism of synthesized images, while providing limited explicit control over whether the generated vessels are located at anatomically consistent positions. However, accurate vascular localization is essential for interventional guidance, because visually plausible but spatially misplaced vessels may provide misleading information for instrument navigation. \textbf{\textit{(ii)} Inefficient modeling of redundant background content.} The objective of this task is not to regenerate the entire image, but to synthesize vascular appearance within anatomically relevant regions. The remaining non-vascular background can largely be preserved from the input non-contrast image. Since the background occupies the majority of pixels, directly modeling the whole image may allocate substantial model capacity to redundant background reconstruction and may weaken the learning of vascular appearance.

To address these challenges, we propose a two-stage \underline{\textbf{Ve}}ssel-focused \underline{\textbf{C}}ontrast-free \underline{\textbf{A}}ngiogram \underline{\textbf{S}}ynthesis (\textbf{VeCAS}) framework. The core idea of VeCAS is to decouple vascular structure localization from angiographic appearance synthesis. \textbf{In Stage I, vascular structures are localized from non-contrast X-ray images.} Specifically, we formulate vascular structure localization as a discriminative problem and predict binary vessel masks, rather than relying on a generative model to implicitly determine vessel locations. This formulation provides explicit spatial supervision for vascular localization. Furthermore, we introduce a cross-modality latent distillation strategy that uses X-ray angiograms during training to transfer vessel-sensitive knowledge to the non-contrast image encoder, thereby improving vascular localization under contrast-free conditions. \textbf{In Stage II, angiographic appearance is synthesized under the guidance of the predicted vessel mask and the non-contrast X-ray image.} We develop a vessel-focused diffusion inpainting model that selectively modifies vascular regions while preserving the non-vascular background from the input image. In addition, a vessel-focused loss is introduced to restrict the learning objective to vascular regions, encouraging the model to concentrate on vascular content rather than unnecessary background reconstruction.

To evaluate the potential utility of VeCAS, we conduct experiments on an in-house dataset collected from patients undergoing lower-limb vascular interventions. In addition to quantitative metrics, we perform visual Turing tests, physician assessments, and phantom-based robotic navigation experiments. These evaluations assess the visual fidelity of the synthesized X-ray angiograms and provide a preliminary assessment of their potential to support interventional navigation.

Our main contributions are as follows:
\begin{itemize}
\item We propose \textbf{VeCAS}, a two-stage vessel-focused framework for contrast-free angiogram synthesis, providing a computational strategy to generate an angiogram-like vascular visualization from non-contrast X-ray images.
\item A cross-modality latent distillation strategy is introduced to transfer vessel-sensitive knowledge from X-ray angiograms to non-contrast X-ray images, improving vascular localization under contrast-free conditions.
\item A vessel-focused inpainting method is developed to synthesize angiographic appearance only within vascular regions while preserving the non-vascular background, thereby reducing unnecessary image modification.
\item VeCAS outperforms existing methods in vascular structural fidelity and image quality. Reader studies further indicate that the synthesized angiograms exhibit perceptual realism, while phantom navigation experiments demonstrate their practical utility, with navigation time and operation steps reduced by $41.4\%$ and $40.7\%$, respectively, compared with non-contrast guidance.
\end{itemize}

The remainder of this paper is organized as follows. Section~\ref{sec:related_work} reviews related work. Section~\ref{sec:method} describes the proposed method in detail. Section~\ref{sec:results} presents the experimental setup and results. Finally, Section~\ref{sec:conclusion} concludes the paper.

%% file: Figure/Figure_Motivation.tex
\begin{figure}[htbp]
\centering\centerline{\includegraphics{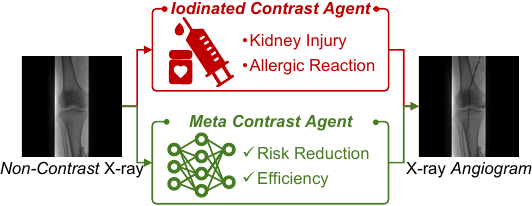}}
\caption{\textbf{Iodinated Contrast Agent \textit{vs}. Meta Contrast Agent.} \textbf{Top:} In conventional X-ray angiography, iodinated contrast agents are injected to enhance vessel visibility, but their use may introduce contrast-related risks, such as kidney injury and allergic reactions. \textbf{Bottom:} The proposed meta contrast agent uses a trained deep neural network to synthesize X-ray angiograms from non-contrast X-ray images, providing a computational alternative with the potential to reduce contrast administration and improve procedural efficiency.}
\label{fig:motivation}
\end{figure}

%% file: Section2_Related_Work.tex
In this section, we review studies related to this work from three perspectives: low-contrast and contrast-free interventions (Section~\ref{sec:related_works_1}), medical image translation (Section~\ref{sec:related_works_2}), and angiogram synthesis (Section~\ref{sec:related_works_3}).

\input{Figure/Figure_VeCAS}

\subsection{Low-Contrast \& Contrast-Free Interventions}
\label{sec:related_works_1}

Reducing contrast administration has become an important goal in image-guided vascular interventions, especially for patients with chronic kidney disease or other contrast-related risk factors~\cite{almendarez2019procedural}. Several clinical strategies have been explored to reduce or avoid iodinated contrast administration. In coronary interventions, intravascular ultrasound- or physiology-guided ultra-low-contrast and zero-contrast procedures have shown feasibility in patients with renal dysfunction~\cite{ali2016imaging,shibata2022feasibility,truong2025ultra}. Roadmap-based guidance has also been investigated to reduce the need for repeated contrast injections by overlaying previously acquired vascular information onto live fluoroscopy~\cite{hennessey2024dynamic}. In peripheral vascular interventions, ${\rm CO}_2$ angiography has been used as an alternative to reduce the use of iodinated contrast agents in high-risk patients~\cite{lee2023carbon,wawer2025safety}. Although these strategies have demonstrated clinical value, they remain limited by their dependence on additional intravascular devices, physiological measurements, alternative contrast agents, or previously acquired angiographic roadmaps. This limitation motivates the development of computational approaches that can synthesize X-ray angiograms directly from non-contrast X-ray images.

\subsection{Medical Image Translation}
\label{sec:related_works_2}

Medical image translation aims to transform images from one imaging domain to another and provides an important technical foundation for angiogram synthesis. Early methods were mainly based on generative adversarial networks (GANs) adapted from natural image translation~\cite{isola2017image,zhu2017unpaired,park2020contrastive}. These frameworks have been further extended to medical imaging by incorporating deformable registration~\cite{kong2021breaking} or improved network architectures~\cite{dalmaz2022resvit} to better handle anatomical variations and modality differences. More recently, diffusion models have shown strong generative modeling capability and have been increasingly adopted for image synthesis tasks~\cite{ho2020denoising,dhariwal2021diffusion}. Diffusion bridge models are particularly relevant to medical image translation because they explicitly model the transformation between two endpoint distributions~\cite{zhou2024denoising}, with representative methods including BBDM~\cite{li2023bbdm} and SelfRDB~\cite{arslan2025self}. Another line of work adapts generative foundation models pre-trained on large-scale natural images~\cite{rombach2022high}, such as ControlNet~\cite{zhang2023adding} and T$2$I-Adapter~\cite{mou2024t2i}, to conditional image translation tasks. Although these methods have achieved promising results, they usually learn a global mapping between image domains and couple vascular localization with angiographic appearance synthesis within a single model. This may result in inaccurate vessel localization and unnecessary background reconstruction, which are particularly undesirable for contrast-free angiogram synthesis. In contrast, VeCAS explicitly decouples localization and synthesis by first predicting a binary vessel mask and then performing vessel-focused inpainting.

\subsection{Angiogram Synthesis}

\label{sec:related_works_3}

Angiogram synthesis has attracted increasing attention because of its potential to reduce the reliance on contrast agents in clinical practice. Existing studies have mainly focused on synthesizing computed tomography angiography (CTA) or magnetic resonance angiography (MRA) from non-contrast CT or MRI using generative models~\cite{xia2023virtual,koch2025cross}. For example, Lyu \textit{et al.}~\cite{lyu2023generative} proposed a GAN-based method for synthesizing CTA from non-contrast CT and demonstrated its potential for vascular disease assessment. Recent diffusion-based methods have further improved vascular structure modeling in angiogram synthesis. Wang \textit{et al.}~\cite{wang2025vastsd} introduced a $3$D vascular tree state-space diffusion model to better capture vascular geometry in non-contrast $3$D volumes, while Li \textit{et al.}~\cite{li2025angiodiff} proposed AngioDiff to improve vascular structure preservation and $3$D consistency in CTA synthesis. Several studies have also explored generative models for synthesizing X-ray angiograms from non-contrast X-ray images. CAS-GAN~\cite{huang2025cas} introduced a disentangled representation learning framework to model vessel and background features for contrast-free angiogram synthesis. Wang \textit{et al.}~\cite{wang2025angio} proposed an adversarial diffusion framework that incorporates a parametric vascular model and mask-guided adversarial learning to improve vascular geometric fidelity. However, these methods were mainly developed for coronary angiogram synthesis, where paired training data are difficult to obtain due to substantial cardiac and respiratory motion. In contrast, VeCAS focuses on peripheral angiogram synthesis, where motion-induced deformation is typically less pronounced, making paired learning more feasible.

%% file: Figure/Figure_VeCAS.tex
\begin{figure*}[t]
\centering\centerline{\includegraphics{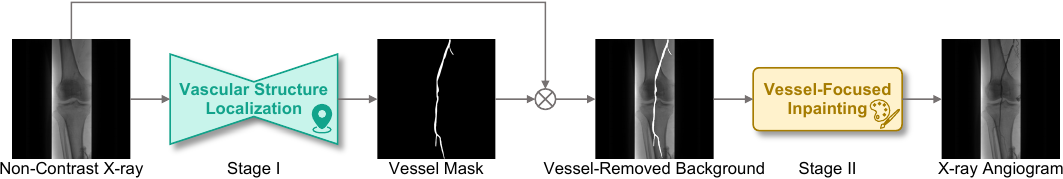}}
\caption{\textbf{Overview of VeCAS.} VeCAS performs contrast-free angiogram synthesis using a two-stage framework. 
\textbf{Stage I:} A vascular structure localization model predicts the vessel mask from a non-contrast X-ray image. 
\textbf{Stage II:} A vessel-focused inpainting model synthesizes angiographic appearance within the predicted vascular regions while preserving the non-vascular background from the input image.}
\label{fig:vecas}
\end{figure*}

%% file: Section3_Method.tex
\subsection{Overview}
VeCAS is a two-stage framework for contrast-free angiogram synthesis, as shown in Fig.~\ref{fig:vecas}. Given a non-contrast X-ray image, VeCAS aims to synthesize angiogram-like vascular appearance while preserving the original non-vascular background. The key idea is to decouple vascular structure localization from angiographic appearance synthesis. In Stage I (Section~\ref{sec:method_stage_i}), a discriminative model predicts a binary vessel mask $\hat{m}\in\{0,1\}^{1\times H\times W}$ from a non-contrast X-ray image $x_{\rm N}\in\mathbb{R}^{C\times H\times W}$, where $C$, $H$, and $W$ denote the number of channels, height, and width, respectively. To exploit the complementary vascular information provided by the corresponding X-ray angiogram $x_{\rm A}\in\mathbb{R}^{C\times H\times W}$ during training, we introduce a cross-modality latent distillation strategy based on feature prediction. In Stage II (Section~\ref{sec:method_stage_ii}), the predicted vessel mask $\hat{m}$ is used to guide a vessel-focused inpainting model to synthesize the X-ray angiogram $\hat{x}_{\rm A}\in\mathbb{R}^{C\times H\times W}$. Unlike conventional diffusion-based methods that model the entire image, VeCAS focuses on vascular regions during both training and inference. Specifically, the vessel-focused loss encourages the model to learn angiographic appearance within vascular regions, while the vessel-focused sampling strategy updates only the vascular regions and preserves the background from the input non-contrast X-ray image.

\subsection{Stage I: Vascular Structure Localization}
\label{sec:method_stage_i}

In this stage, we localize vascular structures from a non-contrast X-ray image $x_{\rm N}$ by predicting a binary vessel mask $\hat{m}$. This step is formulated as a discriminative task rather than a generative modeling problem. This formulation provides explicit spatial supervision for vascular regions, thereby supporting anatomically consistent vessel localization. Given a non-contrast X-ray image, the vessel mask is obtained as:
\begin{align}
    \hat{m}=\mathcal{D}\left(\mathcal{E}\left(x_{\rm N}\right)\right)
\end{align}
where $\mathcal{E}(\cdot)$ and $\mathcal{D}(\cdot)$ denote the encoder and decoder of the segmentation model, respectively.

\textbf{Cross-Modality Latent Distillation.} Learning vascular structures directly from non-contrast X-ray images is challenging because vessel cues are often weak and ambiguous without contrast enhancement. Cross-modality knowledge distillation~\cite{hu2020knowledge,chen2022learning,wang2023prototype} provides a natural way to exploit vascular priors from X-ray angiograms for improved localization. However, conventional strategies may impose overly restrictive supervision in this setting, since vessels are clearly highlighted in angiograms but may be barely visible in the corresponding non-contrast X-ray images. Under such a substantial modality gap, directly aligning features or logits across modalities may hinder optimization rather than facilitate knowledge transfer. To address this issue, we transfer vascular knowledge through latent feature prediction instead of direct feature or logit alignment, making the distillation process more flexible for vascular localization, as shown in Fig.~\ref{fig:stage1}.

\input{Figure/Figure_Stage_I}

Specifically, we first train an angiogram-domain segmentation model $\mathcal{D}_{\rm A}\left(\mathcal{E}_{\rm A}(\cdot)\right)$ using X-ray angiograms $x_{\rm A}$ and their corresponding vessel masks $m$. Since this model is trained on contrast-enhanced images, it learns vessel-sensitive latent representations. We then freeze its parameters and train a dedicated encoder $\mathcal{E}_{\rm N}(\cdot)$ for non-contrast X-ray images. Following U-Net~\cite{ronneberger2015u}, $\mathcal{E}_{\rm A}(\cdot)$ and $\mathcal{E}_{\rm N}(\cdot)$ share the same architecture and extract multi-scale features from paired inputs $(x_{\rm A},x_{\rm N})$:
\begin{align}
    f_{\rm A}^{(i)}=\mathcal{E}_{\rm A}^{(i)}(x_{\rm A}), \quad
    f_{\rm N}^{(i)}=\mathcal{E}_{\rm N}^{(i)}(x_{\rm N})
\end{align}
where $i=1,2,\cdots,5$ denotes the feature scale.

To bridge the modality gap between contrast-enhanced and non-contrast images, we introduce a multi-scale predictor $\mathcal{P}(\cdot)$ to estimate angiogram-domain latent features from non-contrast features:
\begin{align}
    \hat{f}_{\rm A}^{(i)}=\mathcal{P}^{(i)}\left(f_{\rm N}^{(i)}\right)
\end{align}
where $\mathcal{P}^{(i)}(\cdot)$ consists of two consecutive ``Conv-BN-ReLU''. Since deeper features usually encode higher-level semantic information~\cite{long2015fully,lin2017feature}, the predictor is applied only to high-level feature maps, \textit{i.e.}, $i=3,4,5$. An ablation study on this design is provided in the \textbf{\textit{\href{https://dxhuang-casia.github.io/data/vecas_supplementary_material.pdf}{Supplementary Material}}} Table S2.

The predicted multi-scale features are then fed into the frozen angiogram-domain decoder $\mathcal{D}_{\rm A}(\cdot)$ to obtain the vessel mask:
\begin{align}
    \hat{m}=\mathcal{D}_{\rm A}\left(\{\tilde{f}_{\rm A}^{(i)}\}_{i=1}^{5}\right)
\end{align}
where
\begin{align}
    \tilde{f}_{\rm A}^{(i)}=
    \begin{cases}
    f_{\rm N}^{(i)}, & i=1,2 \\
    \hat{f}_{\rm A}^{(i)}, & i=3,4,5
    \end{cases}
\end{align}

To supervise the latent prediction process, we define the prediction loss as:
\begin{align}
    \mathcal{L}_{\rm pred}=\sum_{i=3}^{5}\left\|\hat{f}^{(i)}_{\rm A}-f^{(i)}_{\rm A}\right\|_2^2
\end{align}

\input{Figure/Figure_Stage_II}

In addition, the predicted vessel mask is constrained by a segmentation loss combining binary cross-entropy loss and Dice loss:
\begin{align}
    \mathcal{L}_{\rm seg}=0.5\mathcal{L}_{\rm BCE}+0.5\mathcal{L}_{\rm Dice}
\end{align}

The overall objective of Stage I is formulated as:

\begin{align}
    \mathcal{L}_{\rm distill}=\mathcal{L}_{\rm seg}+\lambda_{\rm pred}\mathcal{L}_{\rm pred}
\end{align}
where $\lambda_{\rm pred}$ controls the trade-off between segmentation supervision and latent prediction. A detailed analysis of $\lambda_{\rm pred}$ is reported in the \textbf{\textit{\href{https://dxhuang-casia.github.io/data/vecas_supplementary_material.pdf}{Supplementary Material}}} Fig. S6.

\textbf{Inference.} During inference, only the non-contrast X-ray image $x_{\rm N}$ is available. It is first passed through the non-contrast encoder $\mathcal{E}_{\rm N}(\cdot)$ to extract multi-scale features. The high-level features are then transformed by the predictor $\mathcal{P}(\cdot)$ into angiogram-domain latent features. Finally, the vessel mask $\hat{m}$ is obtained using the angiogram-domain decoder $\mathcal{D}_{\rm A}(\cdot)$ and subsequently used as the structural condition for Stage II.

\subsection{Stage II: Vessel-Focused Inpainting}
\label{sec:method_stage_ii}

The vessel-focused inpainting model is built on diffusion models~\cite{sohl2015deep,ho2020denoising}, as illustrated in Fig.~\ref{fig:stage2}. The core idea is to learn a vessel-aware reverse diffusion process that synthesizes angiographic appearance within vascular regions while preserving the non-vascular background.

\textbf{Preliminaries.} Diffusion models learn the reverse of a predefined forward noising process, which is formulated as a Markov chain of length $T$. Given a clean image $x_0=x$, the forward process progressively perturbs the image by adding Gaussian noise over $T$ timesteps:
\begin{align}
x_t = \sqrt{\bar{\alpha}_t} x_0 + \sqrt{1 - \bar{\alpha}_t}\,\epsilon, \quad \epsilon \sim \mathcal{N}(\mathbf{0}, I)
\end{align}
where $t \in \{1, \cdots, T\}$ denotes the diffusion timestep, $\alpha_t = 1 - \beta_t$, and $\bar{\alpha}_t = \prod_{i=1}^{t} \alpha_i$. Here, $\beta_t$ follows a predefined variance schedule as in DDPM~\cite{ho2020denoising}.

The reverse process aims to recover $x_0$ from the noisy sample $x_t$. In practice, a neural network $\epsilon_\theta(\cdot)$ is trained to estimate the noise $\epsilon$ conditioned on $x_t$ and timestep $t$.

\textbf{Vessel-Focused Denoising Loss.} The standard diffusion objective supervises noise prediction over the entire image. However, for this task, the relevant content is concentrated in vascular regions. To encourage the model to focus on angiographic appearance, we introduce a vessel-focused denoising loss that restricts supervision to the vessel mask:
\begin{align}
\mathcal{L}_{\rm denoise}=\mathbb{E}_{x_{\rm A},t,\epsilon}\left[\left\|m \odot \left(\epsilon_\theta(x_{{\rm A}, t}, t) -\epsilon\right)\right\|_2^2\right]
\end{align}
where $\odot$ denotes element-wise multiplication.

\textbf{Vessel-Focused Sampling.} During inference, we follow diffusion-based inpainting strategies~\cite{lugmayr2022repaint,zhang2025lefusion} and decouple vessel synthesis from background preservation. Starting from Gaussian noise $x_T \sim \mathcal{N}(\mathbf{0}, I)$, the model iteratively performs reverse denoising to synthesize vessel appearance within the masked region while preserving the background from the input non-contrast X-ray image. At each step, the intermediate denoised result is computed as:
\begin{align}
o_{t-1}
=
\frac{1}{\sqrt{\alpha_t}}
\left(
\hat{x}_t
-
\frac{1 - \alpha_t}{\sqrt{1 - \bar{\alpha}_t}}
\epsilon_\theta(\hat{x}_t, t)
\right)
+
\sigma_t z
\end{align}
where $\sigma_t^2=\frac{1-\bar{\alpha}_{t-1}}{1-\bar{\alpha}_t}\beta_t$ is the variance of the posterior Gaussian distribution $q\left(x_{t-1}\vert x_t,x_0\right)$, and $z\sim\mathcal{N}(\mathbf{0},I)$.

To preserve the original background while synthesizing only vascular regions, we fuse the denoised output with the input non-contrast image according to the vessel mask:
\begin{align}
\hat{x}_{t-1}=o_{t-1} \odot \hat{m}+x_{{\rm N}, t-1} \odot (1 - \hat{m})
\end{align}
where $o_{t-1}$ provides the synthesized vascular appearance, and $x_{{\rm N},t-1}$ denotes the forward-diffused non-contrast X-ray image at timestep $t-1$, which preserves the non-vascular background at the same noise level as the reverse diffusion process. After the final reverse step, the synthesized X-ray angiogram is obtained as $\hat{x}_{\rm A} = \hat{x}_0$.

\textbf{Training-Inference Input Discrepancy.} During training, the diffusion model is supervised using X-ray angiograms $x_{\rm A}$ and their corresponding vessel masks $m$. At inference time, ground-truth masks are unavailable, and the model instead uses the predicted mask $\hat{m}$ from Stage I together with the non-contrast X-ray image $x_{\rm N}$. We further analyze the impact of mask quality on synthesis performance in Section~\ref{sec:ablation}.

%% file: Figure/Figure_Stage_I.tex
\begin{figure}[htbp]
\centering\centerline{\includegraphics{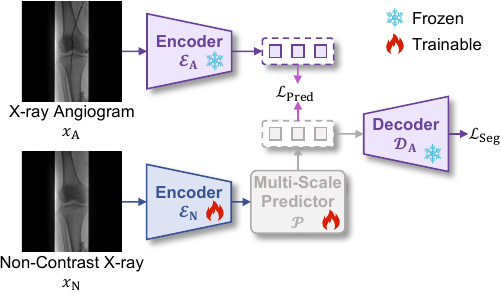}}
\caption{\textbf{Stage I: Cross-Modality Latent Distillation for Vascular Structure Localization.} A frozen angiogram-domain segmentation model $\mathcal{D}_{\rm A}\left(\mathcal{E}_{\rm A}(\cdot)\right)$ provides vessel-sensitive multi-scale features from X-ray angiograms. The trainable non-contrast encoder $\mathcal{E}_{\rm N}(\cdot)$ and multi-scale predictor $\mathcal{P}(\cdot)$ learn to predict angiogram-domain latent features from non-contrast X-ray images. The predicted features are then decoded by the frozen decoder $\mathcal{D}_{\rm A}(\cdot)$ for vessel mask prediction.}
\label{fig:stage1}
\end{figure}

%% file: Figure/Figure_Stage_II.tex
\begin{figure*}[t]
\centering\centerline{\includegraphics{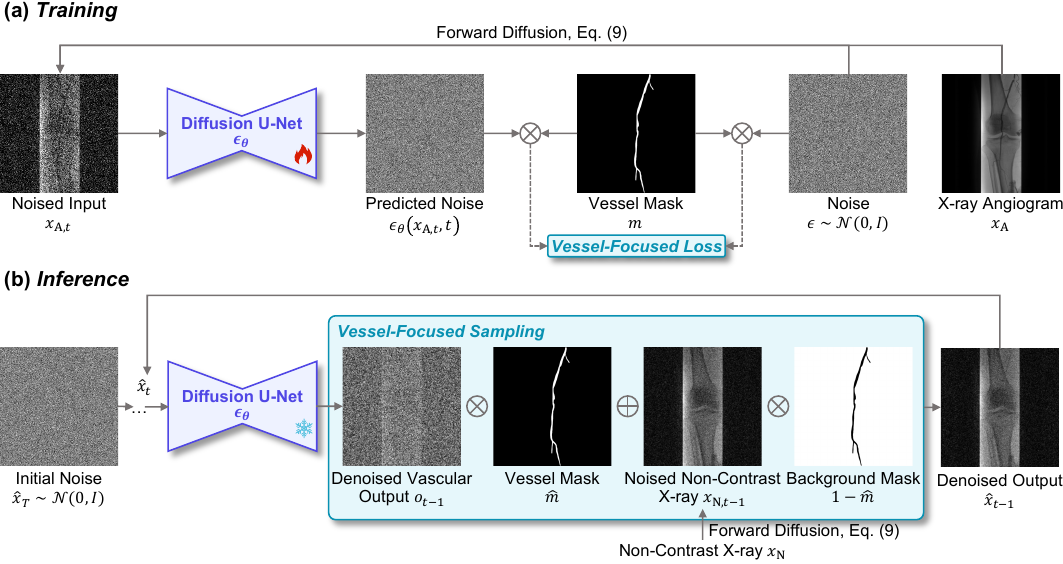}}
\caption{\textbf{Stage II: Vessel-Focused Inpainting.} \textbf{(a) Training:} A vessel-focused denoising loss restricts the diffusion learning objective to vascular regions, encouraging the model to learn angiographic vessel appearance rather than redundant background reconstruction. \textbf{(b) Inference:} At each reverse diffusion step, the denoised vascular output is fused with the forward-diffused non-contrast X-ray image according to the predicted vessel mask from Stage I, thereby synthesizing contrast-enhanced appearance while preserving the non-vascular background.}
\label{fig:stage2}
\end{figure*}

%% file: Section4_Results.tex
\subsection{Clinical Dataset}
\label{sec:dataset}

The dataset used in this study was collected at Beijing Friendship Hospital, Capital Medical University\footnote{\href{https://www.bfh.com.cn/Html/Index.html}{https://www.bfh.com.cn/Html/Index.html}}. It consists of $599$ sequences from $102$ patients who underwent lower-limb vascular interventions. For each sequence, one pre-contrast frame and one contrast-enhanced frame with clear vessel opacification were manually selected from the same sequence. Given the limited motion in lower-limb acquisitions, frame pairs without evident spatial displacement were treated as approximately aligned. This yielded $599$ pairs of non-contrast X-ray images and corresponding X-ray angiograms. The vascular regions in all X-ray angiograms were first annotated by a physician using LabelMe~\cite{russell2008labelme}, and the annotations were then reviewed and corrected by a senior physician.

To evaluate VeCAS, we performed five-fold cross-validation at the patient level. In each fold, $80\%$ of the patients were used for training, and the remaining $20\%$ were used for testing. All sequences from the same patient were assigned to the same subset to avoid patient-level data leakage. The final performance is reported as ``${\rm mean}\pm{\rm std}$'' across the five folds.

\textbf{Ethics Approval.} This study was approved by the Ethics Committee of Beijing Friendship Hospital, Capital Medical University (Approval No. 2020-P2-073-02). Written informed consent was obtained from all participants before data collection. All images were de-identified during preprocessing, and all procedures were conducted in accordance with the Declaration of Helsinki.

\subsection{Implementation Details}
\label{sec:implementation}

All experiments were implemented using PyTorch ($2.8.0$) and Python ($3.11.13$) under Ubuntu ($20.04.3$). The input image size was set to $256\times256$ for all experiments.

\textbf{Stage I.} We adopted U-Net~\cite{ronneberger2015u} as the backbone for vascular structure localization. The angiogram-domain segmentation model $\mathcal{D}_{\rm A}\left(\mathcal{E}_{\rm A}(\cdot)\right)$ was trained for $100$ epochs using AdamW~\cite{loshchilov2019decoupled} with a learning rate of $3\times10^{-4}$. Subsequently, the angiogram-domain model was frozen, and the non-contrast encoder $\mathcal{E}_{\rm N}(\cdot)$ together with the multi-scale predictor $\mathcal{P}(\cdot)$ was trained for $100$ epochs using AdamW with a learning rate of $4\times10^{-4}$. The batch size was set to $16$. Training was performed on a single NVIDIA RTX $4090$ GPU with $24$ GB of memory.

\textbf{Stage II.} The vessel-focused inpainting model was implemented using the Diffusers\footnote{\href{https://huggingface.co/docs/diffusers/index}{https://huggingface.co/docs/diffusers/index}} library and operated in image space. The model was optimized using AdamW with a learning rate of $1\times10^{-4}$ and a batch size of $8$ for $300$ epochs. Training was performed on a single NVIDIA RTX $4090$ GPU with $24$ GB of memory. During inference, the number of sampling steps was set to $100$. The influence of the number of sampling steps is analyzed in the \textbf{\textit{\href{https://dxhuang-casia.github.io/data/vecas_supplementary_material.pdf}{Supplementary Material}}} Table S4.

\input{Table/Table_Visual_Turing_Test}

\textbf{Evaluation Metrics.} For Stage I, we report the Dice similarity coefficient (DSC) and centerline Dice (clDice)~\cite{shit2021cldice}. DSC measures the pixel-level overlap between the predicted and ground-truth vessel masks, while clDice better reflects the topological consistency of tubular vascular structures. For Stage II, DSC and clDice are also used to evaluate the vascular structural fidelity of synthesized X-ray angiograms. Specifically, the angiogram-domain segmentation model $\mathcal{D}_{\rm A}\left(\mathcal{E}_{\rm A}(\cdot)\right)$, trained on real X-ray angiograms, is used to extract vessel masks from synthesized angiograms. In addition, we report two commonly used image-quality metrics, including peak signal-to-noise ratio (PSNR) and structural similarity index measure (SSIM), to evaluate whole-image fidelity.

\subsection{Perceptual and Physician Assessment}
\label{sec:physician_assessment}
\textbf{Visual Turing Test.} To assess the perceptual realism of the synthesized X-ray angiograms, we conducted a Visual Turing Test~\cite{geman2015visual}. Four independent readers with different levels of expertise evaluated $100$ images, including $50$ real X-ray angiograms and $50$ synthesized X-ray angiograms, in a randomized and blinded setting. As reported in Table~\ref{table:turing}, the readers achieved an average Sen. of $77.50\%$, Spe. of $60.50\%$, and Acc. of $69.00\%$. Because synthesized images were treated as negative samples, the relatively low Spe. indicates that readers often classified them as real, suggesting that some synthesized angiograms were visually difficult to distinguish from real angiograms.

\textbf{Physician Assessment.} To further assess the potential procedural usefulness of the synthesized angiograms, three experienced physicians independently evaluated $50$ synthesized angiograms by comparing them with the corresponding real angiograms. The assessment used three criteria: vascular morphological realism (VMR), contrast opacification realism (COR), and procedural usability (PU).
\begin{itemize}
\item \textbf{Vascular Morphological Realism} evaluates whether the synthesized angiograms are consistent with the real angiograms in terms of vessel locations and trajectories, continuity, boundary definition, and major branches.
\item \textbf{Contrast Opacification Realism} assesses the naturalness of the synthesized contrast appearance with respect to intravascular intensity distribution, contrast variation, and edge transitions.
\item \textbf{Procedural Usability} measures whether the synthesized angiograms provide useful information about vessel trajectories and major branches for potential procedural guidance.
\end{itemize}

\input{Figure/Figure_Clinical_Scores}

A five-point Likert scale was used for each criterion, with higher scores indicating better perceived quality. Detailed scoring rules are provided in the \textbf{\textit{\href{https://dxhuang-casia.github.io/data/vecas_supplementary_material.pdf}{Supplementary Material}}} Section S1. As shown in Fig.~\ref{fig:clinical_scores}, the synthesized angiograms received average scores above $3.00$ from all three physicians across the three criteria. The pairwise linearly weighted Cohen's $\kappa$ coefficients ranged from $0.46$ to $0.84$, indicating moderate to high inter-physician agreement. Overall, these results suggest that the synthesized angiograms achieve generally acceptable quality and may provide useful angiographic visualization from non-contrast X-ray images.

\subsection{Main Results}
\label{sec:main_results}

\textbf{Stage I.} Table~\ref{table:stage_i} reports the quantitative results of vascular structure localization. The compared methods include generative models, segmentation models, and cross-modality interaction methods. Generative models, such as Pix$2$Pix~\cite{isola2017image}, CycleGAN~\cite{zhu2017unpaired}, BBDM~\cite{li2023bbdm}, and ControlNet~\cite{zhang2023adding}, show limited performance, indicating that direct image translation is not well suited for localizing vascular structures from non-contrast X-ray images. In contrast, segmentation models achieve substantially better results, supporting our formulation of vascular localization as a discriminative task. Among the compared segmentation models, UNet++~\cite{zhou2020unet++} obtains the best performance. Nevertheless, VeCAS further improves DSC and clDice by $+1.17\%$ and $+0.75\%$, respectively, demonstrating the benefit of cross-modality latent distillation. Compared with existing cross-modality interaction methods, VeCAS also achieves the best overall performance, suggesting that latent feature prediction is more effective than direct feature or logit alignment given the large appearance gap between non-contrast X-ray images and X-ray angiograms. Qualitative results are provided in the \textbf{\textit{\href{https://dxhuang-casia.github.io/data/vecas_supplementary_material.pdf}{Supplementary Material}}} Fig.~S3.

\input{Table/Table_Stage_I}
\input{Table/Table_Stage_I_Plus_II}

\textbf{Stage I $+$ II.} Table~\ref{table:stage2} reports the quantitative comparison of the complete pipeline. VeCAS achieves the best vascular structural fidelity, with a DSC of $42.27\%$ and a clDice of $47.70\%$. Compared with the second-best method for each metric, VeCAS improves DSC by $+5.84\%$ over ControlNet and clDice by $+5.92\%$ over RegGAN~\cite{kong2021breaking}. These results indicate that explicitly decoupling vascular structure localization from angiographic appearance synthesis improves vascular structural fidelity. Several baseline methods obtain competitive PSNR or SSIM, while their DSC and clDice remain much lower than those of VeCAS. This suggests that whole-image metrics are dominated by the large non-vascular background and may not fully reflect vascular structural fidelity. In contrast, VeCAS achieves a better balance between image fidelity and vascular structural fidelity by preserving the background while focusing synthesis on vascular regions. Fig.~\ref{fig:stage_ii_vis} provides qualitative comparisons of different methods.\input{Table/Table_Distill_Loss}Most existing methods preserve the background appearance but generate weak, incomplete, or barely visible vascular structures. By contrast, VeCAS produces clearer and more continuous vessels while maintaining the non-vascular background.

\input{Figure/Figure_Stage_I_Plus_II_Vis}
\input{Figure/Figure_Corrupt}
\input{Table/Table_Vessel_Focused}

\subsection{Ablation Study}
\label{sec:ablation}

\textbf{Distillation Loss $\mathcal{L}_{\rm distill}$.} Table~\ref{table:distill_loss} reports the ablation results of the distillation loss in Stage I. Using only $\mathcal{L}_{\rm pred}$ leads to poor performance, indicating that latent feature prediction alone is insufficient for accurate vascular localization. In contrast, $\mathcal{L}_{\rm seg}$ provides direct pixel-level supervision and substantially improves the results. Combining $\mathcal{L}_{\rm seg}$ and $\mathcal{L}_{\rm pred}$ achieves the best performance, demonstrating that latent prediction provides complementary cross-modality supervision beyond standard segmentation training.

\textbf{Vessel-Focused Design.} Table~\ref{table:vessel} reports the ablation results of the vessel-focused design in Stage II. Removing both the vessel-focused loss and vessel-focused sampling results in poor performance across all metrics. Using only vessel-focused sampling improves PSNR and SSIM, but DSC and clDice remain low, suggesting that this strategy mainly benefits background preservation. In contrast, using only the vessel-focused loss improves DSC and clDice, indicating that emphasizing vascular regions during training is important for learning angiographic appearance. Combining both components yields the best overall performance. These results show that the vessel-focused loss and the vessel-focused sampling play complementary roles. Qualitative results are provided in the \textbf{\textit{\href{https://dxhuang-casia.github.io/data/vecas_supplementary_material.pdf}{Supplementary Material}}} Fig.~S5.

\input{Figure/Figure_Phantom_Experiment_Results}
\textbf{Error Propagation from Stage I to Stage II.} Since VeCAS adopts a two-stage design, we further analyze how mask quality affects the final synthesis results. Stage II is evaluated using the predicted masks from Stage I, the ground-truth masks, and several perturbed ground-truth masks. As shown in Fig.~\ref{fig:corrupt}, using the ground-truth mask provides an oracle upper bound, achieving $60.22\%$ DSC and $66.54\%$ clDice. In comparison, the complete VeCAS pipeline using the predicted mask achieves $42.27\%$ DSC and $47.70\%$ clDice, indicating that mask quality directly affects vascular fidelity. Among different perturbations, random translation causes the largest performance degradation, suggesting that the inpainting model is sensitive to spatial misalignment because the mask determines where angiographic appearance is synthesized. Erosion and dilation also degrade the results, suggesting that accurate vessel boundaries and calibers are important for preserving vascular structural fidelity. In contrast, local vessel breaks and false-positive branches cause less performance degradation, likely because the global vessel location is still largely preserved. PSNR and SSIM remain relatively stable across different inputs, further confirming that whole-image fidelity metrics are dominated by the preserved background and may be insensitive to vascular structural errors.

\subsection{Phantom Experiment}
\label{sec:phantom_experiment}

\textbf{Experimental Setup.} A simulated vascular intervention environment, as shown in Fig.~\ref{fig:phantom_experiment}(a), was constructed to evaluate the feasibility of VeCAS-guided guidewire navigation. The platform consists of a vascular phantom, a vascular robotic system~\cite{zhao2021design}, a guidewire, an overhead camera, and a local workstation. Each vascular phantom was fabricated based on the vessel annotation of a real X-ray angiogram, and the corresponding non-contrast X-ray image was spatially aligned with the phantom. During navigation, the guidewire was manipulated by the robotic system using an Xbox game controller, following the control scheme described in~\cite{li2024casog}. The guidewire was captured in real time, segmented using threshold-based preprocessing, and overlaid on the displayed image to simulate visual feedback during image-guided interventions. For VeCAS-guided navigation, the non-contrast X-ray image was sent from the local workstation to a remote server via SSH, and the synthesized X-ray angiogram was then transferred back and displayed on the local interface. More details are provided in the \textbf{\textit{\href{https://dxhuang-casia.github.io/data/vecas_supplementary_material.pdf}{Supplementary Material}}} Section S2.

\textbf{Protocol.} Three operators were invited to perform guidewire navigation on two vascular phantoms. For each phantom, each operator was asked to navigate the guidewire tip from the inlet to a predefined distal target point. Three guidance conditions were compared: \ding{182} non-contrast guidance, \ding{183} VeCAS guidance, and \ding{184} contrast guidance. For each operator, phantom, and guidance condition, the task was repeated five times.

\textbf{Evaluation Metrics.} Navigation performance was evaluated using two task-level metrics: time to target and the number of operation steps. Time to target was measured from the first control command to the arrival of the guidewire tip at the target region. The number of operation steps was used to quantify manipulation complexity, where one step was defined as one discrete control command, such as guidewire advancement, withdrawal, clockwise rotation, or counterclockwise rotation.

\textbf{Main Results.} As shown in Fig.~\ref{fig:phantom_experiment}(b), VeCAS guidance substantially improved robotic guidewire navigation efficiency compared with non-contrast guidance. Specifically, the time to target was reduced by $41.4\%$, and the number of operation steps was reduced by $40.7\%$. VeCAS guidance also achieved navigation performance comparable to contrast guidance. These results demonstrate the potential of VeCAS to provide contrast-free visual guidance for interventional navigation.

%% file: Table/Table_Visual_Turing_Test.tex
\begin{table}[htbp]
\caption{Results of the Visual Turing Test. Sensitivity (Sen.), specificity (Spe.), and accuracy (Acc.) are reported for each reader.}
\label{table:turing}
\centering
\begin{tabular}{lccccc}
\toprule
Metric & R$1$ & R$2$ & R$3$ & R$4$ & ${\rm mean\pm std}$ \\ \midrule
Sen. (\%) & $62.00$ & $76.00$ & $94.00$ & $78.00$ & $77.50\pm13.10$ \\
Spe. (\%) & $50.00$ & $52.00$ & $62.00$ & $78.00$ & $60.50\pm12.79$ \\
Acc. (\%) & $56.00$ & $64.00$ & $78.00$ & $78.00$ & $69.00\pm10.89$ \\ \bottomrule
 \multicolumn{6}{l}{Positives: Real X-ray angiograms.} \\
 \multicolumn{6}{l}{Negatives: Synthesized X-ray angiograms.} \\
\end{tabular}
\end{table}

%% file: Figure/Figure_Clinical_Scores.tex
\begin{figure}[htbp]
\centering\centerline{\includegraphics{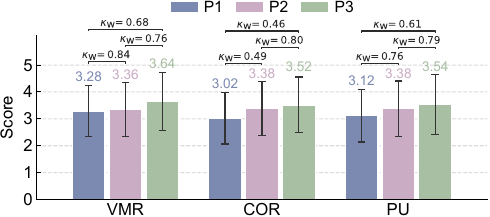}}
\caption{\textbf{Physician Assessments of Synthesized Angiograms.} Three physicians independently scored the synthesized angiograms using three criteria: vascular morphological realism (VMR), contrast opacification realism (COR), and procedural usability (PU).}
\label{fig:clinical_scores}
\end{figure}

%% file: Table/Table_Stage_I.tex
\begin{table}[htbp]
\caption{Quantitative comparison of different methods for Stage I. Results are reported as ``${\rm mean}\pm{\rm std}$'' over five-fold cross-validation. The best and second-best results are highlighted in \textbf{bold} and \underline{underlined}, respectively. Statistical significance against VeCAS is assessed using Wilcoxon signed-rank tests on patient-level scores ($\ast$: $p<0.05$, $\ast\ast$: $p<0.01$, $\ast\ast\ast$: $p<0.001$).}
\label{table:stage_i}
\centering
\begin{tabular}{lll}
\toprule
Method & DSC (\%) $\uparrow$ & clDice (\%) $\uparrow$ \\ \midrule
\multicolumn{3}{l}{\textit{\color{gray}Generative Models}} \\
Pix$2$Pix~\cite{isola2017image} {\tiny\color{gray}[\textit{CVPR}'17]} & ${9.75\pm0.92}^{\ast\ast\ast}$ & ${9.50\pm1.10}^{\ast\ast\ast}$ \\
CycleGAN~\cite{zhu2017unpaired} {\tiny\color{gray}[\textit{ICCV}'17]} & ${2.30\pm2.47}^{\ast\ast\ast}$ & ${2.89\pm3.42}^{\ast\ast\ast}$ \\
ResViT~\cite{dalmaz2022resvit} {\tiny\color{gray}[\textit{TMI}'22]} & ${37.03\pm7.09}^{\ast\ast\ast}$ & ${42.69\pm7.93}^{\ast\ast\ast}$ \\
BBDM~\cite{li2023bbdm} {\tiny\color{gray}[\textit{CVPR}'23]} & ${5.43\pm1.67}^{\ast\ast\ast}$ & ${5.38\pm1.86}^{\ast\ast\ast}$ \\
ControlNet~\cite{zhang2023adding} {\tiny\color{gray}[\textit{ICCV}'23]} & ${8.93\pm0.64}^{\ast\ast\ast}$ & ${8.49\pm0.76}^{\ast\ast\ast}$ \\
\midrule
\multicolumn{3}{l}{\textit{\color{gray}Segmentation Models}} \\
U-Net~\cite{ronneberger2015u} {\tiny\color{gray}[\textit{MICCAI}'15]} & ${51.29\pm2.91}^{\ast\ast}$ & ${59.85\pm3.98}$ \\
Attention U-Net~\cite{schlemper2019attention} {\tiny\color{gray}[\textit{MedIA}'19]} & ${50.98\pm3.06}^{\ast\ast}$ & ${59.49\pm3.84}^\ast$ \\
UNet++~\cite{zhou2020unet++} {\tiny\color{gray}[\textit{TMI}'20]} & $\underline{51.39\pm3.52}^{\ast\ast\ast}$ & ${59.87\pm4.39}^{\ast}$ \\
UCTransNet~\cite{wang2022uctransnet} {\tiny\color{gray}[\textit{AAAI}'22]} & ${51.07\pm3.48}^{\ast\ast}$ & ${59.44\pm4.31}^{\ast}$ \\
VM-UNet~\cite{ruan2025vm} {\tiny\color{gray}[\textit{ACM TOMM}'25]} & ${47.64\pm5.65}^{\ast\ast\ast}$ & ${54.95\pm7.52}^{\ast\ast\ast}$ \\
MambaLiteUNet~\cite{rahman2026mamba} {\tiny\color{gray}[\textit{CVPR}'26]} & ${45.93\pm3.71}^{\ast\ast\ast}$ & ${52.59\pm4.68}^{\ast\ast\ast}$ \\
\midrule
\multicolumn{3}{l}{\textit{\color{gray}Cross-modality Interaction Methods}} \\
Joint Learning & ${50.86\pm3.66}^{\ast\ast}$ & ${58.74\pm4.86}^{\ast\ast}$ \\
KD-Net~\cite{hu2020knowledge} {\tiny\color{gray}[\textit{MICCAI}'20]} & ${51.26\pm2.91}^{\ast\ast}$ & $\underline{60.59\pm4.17}$ \\
PMKL~\cite{chen2022learning} {\tiny\color{gray}[\textit{TMI}'22]} & ${50.60\pm3.79}^{\ast\ast\ast}$ & ${59.03\pm4.86}^{\ast\ast}$ \\
ProtoKD~\cite{wang2023prototype} {\tiny\color{gray}[\textit{ICASSP}'23]} & ${49.92\pm3.09}^{\ast\ast\ast}$ & ${57.92\pm3.91}^{\ast\ast\ast}$ \\
CroDiNo-KD~\cite{ferrod2025revisiting} {\tiny\color{gray}[\textit{ECML-PKDD}'25]} & ${49.04\pm3.13}^{\ast\ast\ast}$ & ${56.84\pm3.75}^{\ast\ast\ast}$ \\
\rowcolor{lightblue} VeCAS (Stage I) {\tiny\color{gray}[\textit{\textbf{Ours}}]} & $\bm{52.56\pm3.32}$ & $\bm{60.62\pm4.42}$ \\
\bottomrule
\end{tabular}
\end{table}

%% file: Table/Table_Stage_I_Plus_II.tex
\begin{table}[htbp]
\caption{Quantitative comparison of different methods for Stage I $+$ II. Results are reported as ``${\rm mean}\pm{\rm std}$'' over five-fold cross-validation. The best and second-best results are highlighted in \textbf{bold} and \underline{underlined}, respectively. Statistical significance against VeCAS is assessed using Wilcoxon signed-rank tests on patient-level scores ($\ast$: $p<0.05$, $\ast\ast$: $p<0.01$, $\ast\ast\ast$: $p<0.001$).}
\label{table:stage2}
\centering
\resizebox{\linewidth}{!}{
\begin{tabular}{lllll}
\toprule
Method & DSC (\%) $\uparrow$ & clDice (\%) $\uparrow$ & PSNR (dB) $\uparrow$ & SSIM $\uparrow$ \\ \midrule
Pix$2$Pix~\cite{isola2017image} {\tiny\color{gray}[\textit{CVPR}'17]} & ${13.11\pm2.26}^{\ast\ast\ast}$ & ${14.78\pm2.46}^{\ast\ast\ast}$ & ${36.06\pm1.20}^{\ast\ast\ast}$ & ${0.9355\pm0.0173}^{\ast\ast\ast}$ \\
CycleGAN~\cite{zhu2017unpaired} {\tiny\color{gray}[\textit{ICCV}'17]} & ${26.41\pm4.07}^{\ast\ast\ast}$ & ${31.93\pm5.05}^{\ast\ast\ast}$ & ${35.71\pm0.74}^{\ast\ast\ast}$ & $\underline{0.9498\pm0.0075}^{\ast\ast\ast}$ \\
CUT~\cite{park2020contrastive} {\tiny\color{gray}[\textit{ECCV}'20]} & ${26.72\pm2.96}^{\ast\ast\ast}$ & ${32.13\pm3.71}^{\ast\ast\ast}$ & ${35.34\pm0.51}^{\ast\ast\ast}$ & ${0.9015\pm0.0574}^{\ast\ast\ast}$ \\
RegGAN~\cite{kong2021breaking} {\tiny\color{gray}[\textit{NeurIPS}'21]} & ${35.60\pm4.75}^{\ast\ast\ast}$ & $\underline{41.78\pm6.33}^{\ast\ast\ast}$ & ${33.27\pm1.16}^{\ast\ast\ast}$ & ${0.9430\pm0.0065}^{\ast\ast\ast}$ \\
ResViT~\cite{dalmaz2022resvit} {\tiny\color{gray}[\textit{TMI}'22]} & ${31.06\pm3.96}^{\ast\ast\ast}$ & ${35.78\pm5.20}^{\ast\ast\ast}$ & ${33.56\pm1.01}^{\ast\ast\ast}$ & ${0.9400\pm0.0080}^{\ast\ast\ast}$ \\
BBDM~\cite{li2023bbdm} {\tiny\color{gray}[\textit{CVPR}'23]} & ${30.80\pm 3.95}^{\ast\ast\ast}$ & ${36.54\pm5.23}^{\ast\ast\ast}$ & $\bm{37.67\pm0.55}$ & ${0.9488\pm0.0201}^{\ast\ast\ast}$ \\
ControlNet~\cite{zhang2023adding} {\tiny\color{gray}[\textit{ICCV}'23]} & $\underline{36.43\pm3.44}^{\ast\ast\ast}$ & ${41.19\pm4.38}^{\ast\ast\ast}$ & ${26.04\pm1.04}^{\ast\ast\ast}$ & ${0.7652\pm0.0511}^{\ast\ast\ast}$ \\
SelfRDB~\cite{arslan2025self} {\tiny\color{gray}[\textit{MedIA}'25]} & ${24.79\pm7.03}^{\ast\ast\ast}$ & ${28.84\pm8.71}^{\ast\ast\ast}$ & ${29.72\pm4.91}^{\ast\ast\ast}$ & ${0.8584\pm0.0422}^{\ast\ast\ast}$ \\
\rowcolor{lightblue}VeCAS (Stage I $+$ II) {\tiny\color{gray}[\textit{\textbf{Ours}}]} & $\bm{42.27\pm4.99}$ & $\bm{47.70\pm6.29}$ & $\underline{36.78\pm0.91}$ & $\bm{0.9537\pm0.0183}$ \\
\bottomrule
\end{tabular}}
\end{table}

%% file: Table/Table_Distill_Loss.tex
\begin{table}[htbp]
\caption{Ablation study of the distillation loss $\mathcal{L}_{\rm distill}$ in Stage I. The best and second-best results are highlighted in \textbf{bold} and \underline{underlined}, respectively.}
\label{table:distill_loss}
\centering
\begin{tabular}{llll}
\toprule
$\mathcal{L}_{\rm seg}$ & $\mathcal{L}_{\rm pred}$ & DSC (\%) $\uparrow$ & clDice (\%) $\uparrow$ \\ \midrule
 {\color{red}\XSolidBrush} & {\color{deepgreen}\Checkmark} & $29.49\pm3.53$ & $35.52\pm2.74$ \\
 {\color{deepgreen}\Checkmark} & {\color{red}\XSolidBrush} & $\underline{51.49\pm3.47}$ & $\underline{59.66\pm4.63}$ \\
\rowcolor{lightblue} {\color{deepgreen}\Checkmark} & {\color{deepgreen}\Checkmark} & $\bm{52.56\pm3.32}$ & $\bm{60.62\pm4.42}$ \\ 
 \bottomrule
\end{tabular}
\end{table}

%% file: Figure/Figure_Stage_I_Plus_II_Vis.tex
\begin{figure*}[t]
\centering\centerline{\includegraphics{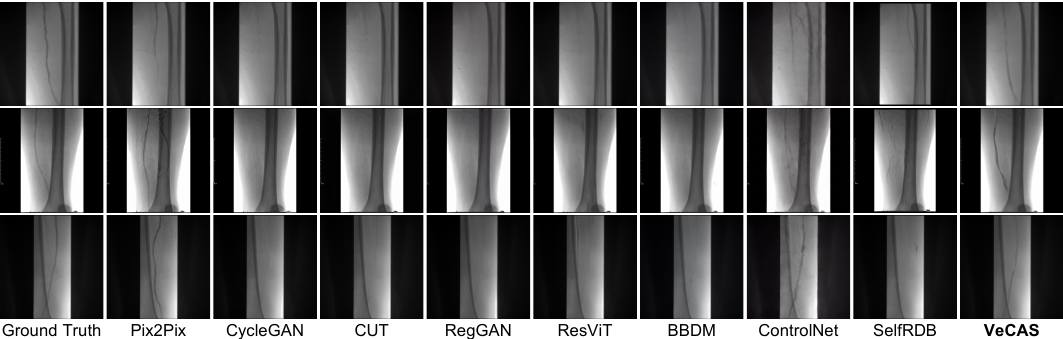}}
\caption{\textbf{Qualitative Results of the Complete Pipeline (Stage I $+$ II).} Compared with baseline methods, VeCAS synthesizes clearer and more continuous vascular structures while preserving the non-vascular background. Most baseline methods retain the background appearance but produce weak, incomplete, or barely visible vessels.}
\label{fig:stage_ii_vis}
\end{figure*}

%% file: Figure/Figure_Corrupt.tex
\begin{figure*}[t]
\centering\centerline{\includegraphics{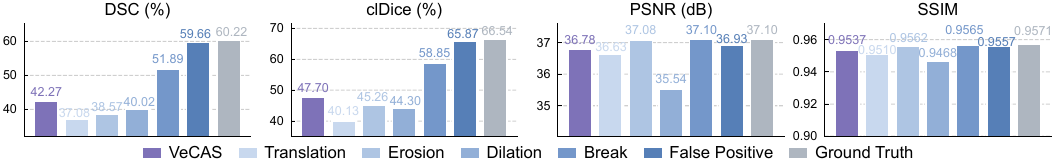}}
\caption{\textbf{Error Propagation Analysis under Different Mask Conditions.} Stage II is evaluated using predicted masks from VeCAS, perturbed ground-truth masks, and oracle ground-truth masks to analyze how mask quality affects performance.}
\label{fig:corrupt}
\end{figure*}

%% file: Table/Table_Vessel_Focused.tex
\begin{table}[htbp]
\caption{Ablation study of the vessel-focused design in Stage II. The best and second-best results are highlighted in \textbf{bold} and \underline{underlined}, respectively.}
\label{table:vessel}
\centering
\resizebox{\linewidth}{!}{
\begin{tabular}{llllll}
\toprule
\multicolumn{2}{c}{Vessel-focused} & \multirow{2}{*}{DSC (\%) $\uparrow$} & \multirow{2}{*}{clDice (\%) $\uparrow$} & \multirow{2}{*}{PSNR (dB) $\uparrow$} & \multirow{2}{*}{SSIM $\uparrow$} \\ \cmidrule{1-2}
Loss & Sampling & & & & \\ \midrule
 {\color{red}\XSolidBrush} & {\color{red}\XSolidBrush} & $11.43\pm3.76$ & $12.55\pm4.23$ & $29.69\pm0.34$ & $0.9257\pm0.0152$ \\
 {\color{red}\XSolidBrush} & {\color{deepgreen}\Checkmark} & $11.38\pm3.03$ & $11.89\pm3.56$ & $\underline{36.19\pm0.93}$ & $\underline{0.9508\pm0.0178}$ \\
 {\color{deepgreen}\Checkmark} & {\color{red}\XSolidBrush} & $\underline{20.51\pm4.32}$ & $\underline{22.96\pm5.08}$ & $31.66\pm0.58$ & $0.9322\pm0.0169$ \\
 \rowcolor{lightblue}{\color{deepgreen}\Checkmark} & {\color{deepgreen}\Checkmark} & $\bm{42.27\pm4.99}$ & $\bm{47.70\pm6.29}$ & $\bm{36.78\pm0.91}$ & $\bm{0.9537\pm0.0183}$ \\
\bottomrule
\end{tabular}
}
\end{table}

%% file: Figure/Figure_Phantom_Experiment_Results.tex
\begin{figure*}[t]
\centering
\begin{tabular}{c@{\hspace{0.1cm}}c}
\raisebox{-0.5\height}{\includegraphics{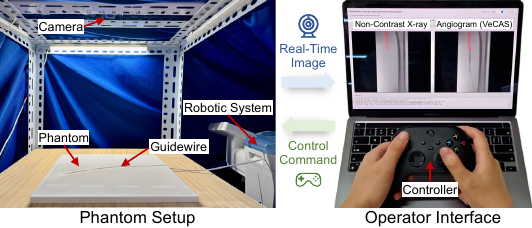}} &
\raisebox{-0.5\height}{\includegraphics{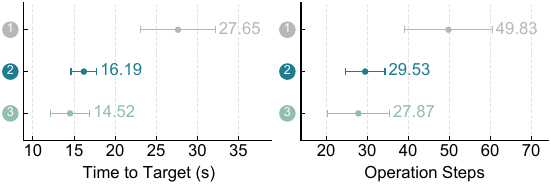}} \\
[1.0mm]
{\fontfamily{phv}\selectfont\fontsize{8pt}{9.5pt}\selectfont
(a) Phantom Experimental Setup} &
{\fontfamily{phv}\selectfont\fontsize{8pt}{9.5pt}\selectfont
(b) Phantom Navigation Results}
\end{tabular}
\caption{\textbf{Phantom Experiment.} (a) Phantom experimental setup. (b) Navigation efficiency under different guidance conditions, evaluated by time to target and the number of operation steps. \ding{182} non-contrast guidance, \ding{183} VeCAS guidance, and \ding{184} contrast guidance.}
\label{fig:phantom_experiment}
\end{figure*}

%% file: Section6_Conclusion.tex
In this paper, we present \textbf{VeCAS}, a \underline{\textbf{Ve}}ssel-focused \underline{\textbf{C}}ontrast-free \underline{\textbf{A}}ngiogram \underline{\textbf{S}}ynthesis framework. VeCAS aims to provide vessel-enhanced visual guidance from non-contrast X-ray images by explicitly decoupling vascular structure localization from angiographic appearance synthesis. Cross-modality latent distillation is used to improve vascular localization by transferring vessel-sensitive knowledge from X-ray angiograms during training. The vessel-focused inpainting model then selectively synthesizes angiographic appearance within the localized vascular regions while preserving the non-vascular background. Quantitative experiments, physician assessments, and robotic guidewire navigation experiments in vascular phantoms collectively support the performance of VeCAS and suggest its potential to reduce the need for iodinated contrast administration in image-guided vascular interventions.

%% file: tmi.bbl
\begin{thebibliography}{10}
\providecommand{\url}[1]{#1}
\csname url@samestyle\endcsname
\providecommand{\newblock}{\relax}
\providecommand{\bibinfo}[2]{#2}
\providecommand{\BIBentrySTDinterwordspacing}{\spaceskip=0pt\relax}
\providecommand{\BIBentryALTinterwordstretchfactor}{4}
\providecommand{\BIBentryALTinterwordspacing}{\spaceskip=\fontdimen2\font plus
\BIBentryALTinterwordstretchfactor\fontdimen3\font minus \fontdimen4\font\relax}
\providecommand{\BIBforeignlanguage}[2]{{%
\expandafter\ifx\csname l@#1\endcsname\relax
\typeout{** WARNING: IEEEtran.bst: No hyphenation pattern has been}%
\typeout{** loaded for the language `#1'. Using the pattern for}%
\typeout{** the default language instead.}%
\else
\language=\csname l@#1\endcsname
\fi
#2}}
\providecommand{\BIBdecl}{\relax}
\BIBdecl

\bibitem{thukkani2015endovascular}
A.~K. Thukkani and S.~Kinlay, ``Endovascular intervention for peripheral artery disease,'' \emph{Circ. Res.}, vol. 116, no.~9, pp. 1599--1613, 2015.

\bibitem{hiramoto2018interventions}
J.~S. Hiramoto, M.~Teraa, G.~J. de~Borst, and M.~S. Conte, ``Interventions for lower extremity peripheral artery disease,'' \emph{Nat. Rev. Cardiol.}, vol.~15, no.~6, pp. 332--350, 2018.

\bibitem{zhao2025large}
H.~{}Zhao \emph{et~al.}, ``Large-scale pretrained frame generative model enables real-time low-dose {DSA} imaging: An {AI} system development and multi-center validation study,'' \emph{Med}, vol.~6, no.~1, p. 100497, 2025.

\bibitem{zhao2026generative}
H.~Zhao \emph{et~al.}, ``Generative {AI}-based low-dose digital subtraction angiography for intra-operative radiation dose reduction: A randomized controlled trial,'' \emph{Nat. Med.}, vol.~32, pp. 288--296, 2026.

\bibitem{clement2018immediate}
O.~Clement \emph{et~al.}, ``Immediate hypersensitivity to contrast agents: The {French} 5-year {CIRTACI} study,'' \emph{Lancet Discov. Sci.}, vol.~1, pp. 51--61, 2018.

\bibitem{fahling2017understanding}
M.~F{\"a}hling, E.~Seeliger, A.~Patzak, and P.~B. Persson, ``Understanding and preventing contrast-induced acute kidney injury,'' \emph{Nat. Rev. Nephrol.}, vol.~13, no.~3, pp. 169--180, 2017.

\bibitem{mehran2006contrast}
R.~Mehran and E.~Nikolsky, ``Contrast-induced nephropathy: Definition, epidemiology, and patients at risk,'' \emph{Kidney Int.}, vol.~69, pp. S11--S15, 2006.

\bibitem{tehrani2013contrast}
S.~Tehrani, C.~Laing, D.~M. Yellon, and D.~J. Hausenloy, ``Contrast-induced acute kidney injury following {PCI},'' \emph{Eur. J. Clin. Invest.}, vol.~43, no.~5, pp. 483--490, 2013.

\bibitem{giacoppo2015impact}
D.~Giacoppo \emph{et~al.}, ``Impact of contrast-induced acute kidney injury after percutaneous coronary intervention on short-and long-term outcomes: Pooled analysis from the {HORIZONS}-{AMI} and {ACUITY} trials,'' \emph{Circ. Cardiovasc. Interv.}, vol.~8, no.~8, p. e002475, 2015.

\bibitem{bartholomew2004impact}
B.~A. Bartholomew \emph{et~al.}, ``Impact of nephropathy after percutaneous coronary intervention and a method for risk stratification,'' \emph{Am. J. Cardiol.}, vol.~93, no.~12, pp. 1515--1519, 2004.

\bibitem{amin2017association}
A.~P. Amin, R.~G. Bach, M.~L. Caruso, K.~F. Kennedy, and J.~A. Spertus, ``Association of variation in contrast volume with acute kidney injury in patients undergoing percutaneous coronary intervention,'' \emph{JAMA Cardiol.}, vol.~2, no.~9, pp. 1007--1012, 2017.

\bibitem{wang2025angio}
Z.~Wang, R.~Yi, X.~Wen, C.~Zhu, K.~Xu, and K.~He, ``{Angio-Diff}: Learning a self-supervised adversarial diffusion model for angiographic geometry generation,'' \emph{The Vis. Comput.}, vol.~41, no.~12, pp. 10\,303--10\,315, 2025.

\bibitem{huang2025cas}
D.-X. Huang \emph{et~al.}, ``{CAS-GAN} for contrast-free angiography synthesis,'' in \emph{Proc. SSCI}, 2025, {D}OI: \href{http://dx.doi.org/10.1109/CISM64958.2025.11060856}{10.1109/CISM64958.2025.11060856}.

\bibitem{ali2016imaging}
Z.~A. Ali \emph{et~al.}, ``Imaging-and physiology-guided percutaneous coronary intervention without contrast administration in advanced renal failure: A feasibility, safety, and outcome study,'' \emph{Eur. Heart J.}, vol.~37, no.~40, pp. 3090--3095, 2016.

\bibitem{shibata2022feasibility}
K.~Shibata \emph{et~al.}, ``Feasibility, safety, and long-term outcomes of zero-contrast percutaneous coronary intervention in patients with chronic kidney disease,'' \emph{Circ. J.}, vol.~86, no.~5, pp. 787--796, 2022.

\bibitem{hennessey2024dynamic}
B.~Hennessey \emph{et~al.}, ``Dynamic coronary roadmap versus standard angiography for percutaneous coronary intervention: The randomised, multicentre {DCR4Contrast} trial,'' \emph{EuroIntervention}, vol.~20, no.~3, p. e198, 2024.

\bibitem{lee2023carbon}
S.-R. Lee \emph{et~al.}, ``Carbon dioxide angiography during peripheral vascular interventions is associated with decreased cardiac and renal complications in patients with chronic kidney disease,'' \emph{J. Vasc. Surg.}, vol.~78, no.~1, pp. 201--208, 2023.

\bibitem{rombach2022high}
R.~Rombach, A.~Blattmann, D.~Lorenz, P.~Esser, and B.~Ommer, ``High-resolution image synthesis with latent diffusion models,'' in \emph{Proc. CVPR}, 2022, pp. 10\,684--10\,695.

\bibitem{brooks2024video}
T.~Brooks \emph{et~al.}, ``Video generation models as world simulators,'' \emph{OpenAI Blog}, 2024.

\bibitem{isola2017image}
P.~Isola, J.-Y. Zhu, T.~Zhou, and A.~A. Efros, ``Image-to-image translation with conditional adversarial networks,'' in \emph{Proc. CVPR}, 2017, pp. 1125--1134.

\bibitem{zhu2017unpaired}
J.-Y. Zhu, T.~Park, P.~Isola, and A.~A. Efros, ``Unpaired image-to-image translation using cycle-consistent adversarial networks,'' in \emph{Proc. ICCV}, 2017, pp. 2223--2232.

\bibitem{almendarez2019procedural}
M.~Almendarez \emph{et~al.}, ``Procedural strategies to reduce the incidence of contrast-induced acute kidney injury during percutaneous coronary intervention,'' \emph{JACC Cardiovasc. Interv.}, vol.~12, no.~19, pp. 1877--1888, 2019.

\bibitem{truong2025ultra}
T.~Truong \emph{et~al.}, ``Ultra-low contrast {IVUS}-guided {PCI} in patients with severe chronic kidney disease: An observational prospective study,'' \emph{Circ. Cardiovasc. Interv.}, vol.~18, no.~2, p. e014854, 2025.

\bibitem{wawer2025safety}
R.~P. Wawer Matos~Reimer \emph{et~al.}, ``Safety and evidence of {CO2} as a vascular contrast agent as an alternative to iodine-based contrast media in vascular procedures: A systematic review by the {ESUR Contrast Medium Safety Committee},'' \emph{Eur. Radiol.}, vol.~35, no.~12, pp. 7680--7687, 2025.

\bibitem{park2020contrastive}
T.~Park, A.~A. Efros, R.~Zhang, and J.-Y. Zhu, ``Contrastive learning for unpaired image-to-image translation,'' in \emph{Proc. ECCV}, 2020, pp. 319--345.

\bibitem{kong2021breaking}
L.~Kong \emph{et~al.}, ``Breaking the dilemma of medical image-to-image translation,'' in \emph{Proc. NeurIPS}, vol.~34, 2021, pp. 1964--1978.

\bibitem{dalmaz2022resvit}
O.~Dalmaz, M.~Yurt, and T.~{\c{C}}ukur, ``{ResViT}: Residual vision transformers for multimodal medical image synthesis,'' \emph{IEEE TMI}, vol.~41, no.~10, pp. 2598--2614, 2022.

\bibitem{ho2020denoising}
J.~Ho, A.~Jain, and P.~Abbeel, ``Denoising diffusion probabilistic models,'' in \emph{Proc. NeurIPS}, vol.~33, 2020, pp. 6840--6851.

\bibitem{dhariwal2021diffusion}
P.~Dhariwal and A.~Nichol, ``Diffusion models beat {GANs} on image synthesis,'' in \emph{Proc. NeurIPS}, vol.~34, 2021, pp. 8780--8794.

\bibitem{zhou2024denoising}
L.~Zhou, A.~Lou, S.~Khanna, and S.~Ermon, ``Denoising diffusion bridge models,'' in \emph{Proc. ICLR}, 2024.

\bibitem{li2023bbdm}
B.~Li, K.~Xue, B.~Liu, and Y.-K. Lai, ``{BBDM}: Image-to-image translation with brownian bridge diffusion models,'' in \emph{Proc. CVPR}, 2023, pp. 1952--1961.

\bibitem{arslan2025self}
F.~Arslan, B.~Kabas, O.~Dalmaz, M.~Ozbey, and T.~{\c{C}}ukur, ``Self-consistent recursive diffusion bridge for medical image translation,'' \emph{MedIA}, vol. 106, p. 103747, 2025.

\bibitem{zhang2023adding}
L.~Zhang, A.~Rao, and M.~Agrawala, ``Adding conditional control to text-to-image diffusion models,'' in \emph{Proc. ICCV}, 2023, pp. 3836--3847.

\bibitem{mou2024t2i}
C.~Mou \emph{et~al.}, ``{T2I-Adapter}: Learning adapters to dig out more controllable ability for text-to-image diffusion models,'' in \emph{Proc. AAAI}, vol.~38, no.~5, 2024, pp. 4296--4304.

\bibitem{xia2023virtual}
Y.~Xia, N.~Ravikumar, T.~Lassila, and A.~F. Frangi, ``Virtual high-resolution {MR} angiography from non-angiographic multi-contrast {MRIs}: Synthetic vascular model populations for in-silico trials,'' \emph{MedIA}, vol.~87, p. 102814, 2023.

\bibitem{koch2025cross}
A.~Koch \emph{et~al.}, ``Cross-modality image synthesis from {TOF-MRA} to {CTA} using diffusion-based models,'' \emph{MedIA}, p. 103722, 2025.

\bibitem{lyu2023generative}
J.~Lyu \emph{et~al.}, ``Generative adversarial network--based noncontrast {CT} angiography for aorta and carotid arteries,'' \emph{Radiology}, vol. 309, no.~2, p. e230681, 2023.

\bibitem{wang2025vastsd}
Z.~Wang, R.~Yi, X.~Wen, C.~Zhu, and K.~Xu, ``{VasTSD}: Learning 3{D} vascular tree-state space diffusion model for angiography synthesis,'' in \emph{Proc. CVPR}, 2025, pp. 15\,693--15\,702.

\bibitem{li2025angiodiff}
A.~Li, W.~Fang, G.~Yang, and M.~Xu, ``{AngioDiff}: Structure-preserving and 3{D}-consistent diffusion for {CT} angiography synthesis,'' in \emph{Proc. BIBM}, 2025, pp. 2429--2434.

\bibitem{hu2020knowledge}
M.~Hu \emph{et~al.}, ``Knowledge distillation from multi-modal to mono-modal segmentation networks,'' in \emph{Proc. MICCAI}, 2020, pp. 772--781.

\bibitem{chen2022learning}
C.~Chen, Q.~Dou, Y.~Jin, Q.~Liu, and P.~A. Heng, ``Learning with privileged multimodal knowledge for unimodal segmentation,'' \emph{IEEE TMI}, vol.~41, no.~3, pp. 621--632, 2022.

\bibitem{wang2023prototype}
S.~Wang, Z.~Yan, D.~Zhang, H.~Wei, Z.~Li, and R.~Li, ``Prototype knowledge distillation for medical segmentation with missing modality,'' in \emph{Proc. ICASSP}, 2023, {D}OI: \href{http://dx.doi.org/10.1109/ICASSP49357.2023.10095014}{10.1109/ICASSP49357.2023.10095014}.

\bibitem{ronneberger2015u}
O.~Ronneberger, P.~Fischer, and T.~Brox, ``U-{N}et: Convolutional networks for biomedical image segmentation,'' in \emph{Proc. MICCAI}, 2015, pp. 234--241.

\bibitem{long2015fully}
J.~Long, E.~Shelhamer, and T.~Darrell, ``Fully convolutional networks for semantic segmentation,'' in \emph{Proc. CVPR}, 2015, pp. 3431--3440.

\bibitem{lin2017feature}
T.-Y. Lin, P.~Doll{\'a}r, R.~Girshick, K.~He, B.~Hariharan, and S.~Belongie, ``Feature pyramid networks for object detection,'' in \emph{Proc. CVPR}, 2017, pp. 2117--2125.

\bibitem{sohl2015deep}
J.~Sohl-Dickstein, E.~Weiss, N.~Maheswaranathan, and S.~Ganguli, ``Deep unsupervised learning using nonequilibrium thermodynamics,'' in \emph{Proc. ICML}, 2015, pp. 2256--2265.

\bibitem{lugmayr2022repaint}
A.~Lugmayr, M.~Danelljan, A.~Romero, F.~Yu, R.~Timofte, and L.~Van~Gool, ``{RePaint}: Inpainting using denoising diffusion probabilistic models,'' in \emph{Proc. CVPR}, 2022, pp. 11\,461--11\,471.

\bibitem{zhang2025lefusion}
H.~Zhang \emph{et~al.}, ``{LeFusion}: Controllable pathology synthesis via lesion-focused diffusion models,'' in \emph{Proc. ICLR}, 2025.

\bibitem{russell2008labelme}
B.~C. Russell, A.~Torralba, K.~P. Murphy, and W.~T. Freeman, ``Label{M}e: A database and web-based tool for image annotation,'' \emph{IJCV}, vol.~77, no.~1, pp. 157--173, 2008.

\bibitem{loshchilov2019decoupled}
I.~Loshchilov and F.~Hutter, ``Decoupled weight decay regularization,'' in \emph{Proc. ICLR}, 2019.

\bibitem{shit2021cldice}
S.~Shit \emph{et~al.}, ``{clDice}-a novel topology-preserving loss function for tubular structure segmentation,'' in \emph{Proc. CVPR}, 2021, pp. 16\,560--16\,569.

\bibitem{geman2015visual}
D.~Geman, S.~Geman, N.~Hallonquist, and L.~Younes, ``Visual {Turing} test for computer vision systems,'' \emph{PNAS}, vol. 112, no.~12, pp. 3618--3623, 2015.

\bibitem{zhou2020unet++}
Z.~Zhou, M.~M.~R. Siddiquee, N.~Tajbakhsh, and J.~Liang, ``U{N}et++: Redesigning skip connections to exploit multiscale features in image segmentation,'' \emph{IEEE TMI}, vol.~39, no.~6, pp. 1856--1867, 2020.

\bibitem{schlemper2019attention}
J.~Schlemper, O.~Oktay, M.~Schaap, M.~Heinrich, B.~Kainz, B.~Glocker, and D.~Rueckert, ``Attention gated networks: Learning to leverage salient regions in medical images,'' \emph{MedIA}, vol.~53, pp. 197--207, 2019.

\bibitem{wang2022uctransnet}
H.~Wang, P.~Cao, J.~Wang, and O.~R. Zaiane, ``{UCTransNet}: Rethinking the skip connections in {U}-{N}et from a channel-wise perspective with transformer,'' in \emph{Proc. AAAI}, vol.~36, no.~3, 2022, pp. 2441--2449.

\bibitem{ruan2025vm}
J.~Ruan, J.~Li, and S.~Xiang, ``{VM-UNet}: Vision {Mamba} {UNet} for medical image segmentation,'' \emph{ACM TOMM}, 2025.

\bibitem{rahman2026mamba}
M.~M. Rahman, S.~K. Jung, and T.~Hammond, ``{MambaLiteUNet}: Cross-gated adaptive feature fusion for robust skin lesion segmentation,'' in \emph{Proc. CVPR}, 2026, p. 8556–8565.

\bibitem{ferrod2025revisiting}
R.~Ferrod, C.~F. Dantas, L.~Di~Caro, and D.~Ienco, ``Revisiting cross-modal knowledge distillation: A disentanglement approach for {RGBD} semantic segmentation,'' in \emph{Proc. ECML-PKDD}, 2025, pp. 492--508.

\bibitem{zhao2021design}
H.-L. Zhao \emph{et~al.}, ``Design and performance evaluation of a novel vascular robotic system for complex percutaneous coronary interventions,'' in \emph{Proc. EMBC}, 2021, pp. 4679--4682.

\bibitem{li2024casog}
H.~Li, X.-H. Zhou, X.-L. Xie, S.-Q. Liu, Z.-Q. Feng, and Z.-G. Hou, ``{CASOG}: Conservative actor--critic with smooth gradient for skill learning in robot-assisted intervention,'' \emph{IEEE TIE}, vol.~71, no.~7, pp. 7722--7731, 2024.

\end{thebibliography}
